\documentclass[11pt]{article}

\usepackage[preprint]{acl}

\usepackage{times}
\usepackage{latexsym}

\usepackage[T1]{fontenc}

\usepackage[utf8]{inputenc}

\usepackage{microtype}

\usepackage{inconsolata}

\usepackage{graphicx}

\usepackage{hyperref}
\usepackage{url}
\usepackage{times}  
\usepackage{helvet}  
\usepackage{courier}  
\usepackage{graphicx} 
\usepackage{wrapfig}
\usepackage{algorithm}
\usepackage{placeins}
\usepackage{amsmath} 
\usepackage{amsfonts} 
\usepackage{amssymb} 
\usepackage{multirow} 
\usepackage{xcolor}
\usepackage{amsthm} 
\usepackage{booktabs, siunitx}
\usepackage{graphicx}
\usepackage{subcaption}
\usepackage[table]{xcolor} 

\newtheorem{assumption}{Assumption}
\usepackage{algorithm}
\usepackage{algpseudocode}
\usepackage{adjustbox}
\usepackage{bm}

\title{Where Do Prompt Perturbations Break Generation? A Segment-Level View of Robustness in LoRA-Tuned Language Models}

\author{Zhuoyun Li, Boxuan Wang, Jinwei Hu, Zhenglin Huang, Qisong He,  \\ {\bf Xinmiao Huang, 
Guangliang Cheng, Xiaowei Huang, Yi Dong} \\
        School of Computer Science and Informatics, University of Liverpool, UK}

\begin{document}
\maketitle
\begin{abstract}
Large language models are sensitive to minor prompt perturbations, yet existing robustness methods usually enforce consistency at the whole-sequence level. This holistic view can hide an important failure mode: a perturbed response may remain globally similar to the clean one while drifting on a critical entity, relation, or conclusion. We introduce S$^2$R$^2$, a segment-level framework for robust LoRA fine-tuning. S$^2$R$^2$ decomposes clean and perturbed generations into semantic segments, aligns them with an optimal-transport objective, and penalises the segments with the largest meaning drift. To connect this output-side objective with model adaptation, we add an adapter-stability regulariser motivated by segment-level attention reallocation, using LoRA norm control as a tractable proxy for limiting perturbation-amplified evidence shifts. A PAC-Bayesian complexity view further explains why controlling adapter growth may support transfer beyond observed perturbations. Experiments on summarisation benchmarks show that S$^2$R$^2$ improves robustness under typographical noise, deletion, synonym replacement, and paraphrasing, while maintaining competitive clean performance and stronger cross-dataset transfer than consistency-based baselines.
\end{abstract}

\section{Introduction}\label{sec:introduction}

Large Language Models (LLMs) have achieved widespread adoption in numerous applications \citep{Hu_Dong_Sun_Huang_2026,wang2025rethinkingmultiagentintelligencelens,hu2025stopreducingresponsibilityllmpowered}, but their reliability is often compromised by minor perturbations to input prompts \citep{huang2024survey,wang2026chainofthoughtlensevaluatingstructured}. These perturbations can arise from typographical errors, paraphrases, word-level substitutions, or adversarially crafted changes, and may lead to unreliable or unsafe outputs \citep{xhonneux2024efficient, 10900215, paulus2025, lin2025single, gan2024reasoning, rauba2024quantifyingperturbationimpactslarge, uncertainty, wang2023largelanguagemodelsreally}. For generation tasks such as summarisation, even a small change in a key entity, relation, or conclusion can alter the meaning of the entire response. This fragility undermines the trustworthy deployment of LLMs, especially in domains where output faithfulness and semantic stability are critical.

Many researchers have focused on bolstering LLM robustness and have proposed well-designed fine-tuning strategies \citep{qiang-etal-2024-prompt, Rdrop, aghajanyan2021better, Zhu2020FreeLB, jiang-etal-2020-smart}. A common thread in these methods is a holistic treatment of the output, for example by minimising the Kullback-Leibler (KL) divergence or related consistency losses over an entire sequence. While such objectives are effective for improving average robustness, they disregard a basic property of language: semantic information is unevenly distributed across an output. Just as a few keywords can define the meaning of a sentence, particular segments of an LLM output may be more critical to its semantic integrity than others. This non-uniform vulnerability is also consistent with observations in robust learning and adversarial fairness, where different groups or components can contribute unevenly to the overall risk \citep{pmlr-v80-agarwal18a, pmlr-v80-hashimoto18a, jin2025enhancing}. By ignoring this asymmetry, holistic methods may hide local failures in which the overall response remains similar while a critical semantic segment drifts \citep{qiang-etal-2024-prompt}.

This motivates a segment-level view of prompt robustness. Instead of asking only whether the full clean and perturbed outputs are similar, we ask where the generation breaks. This distinction is important because perturbations often do not damage all parts of the output uniformly. A response may preserve most of its surface form while changing a medical condition, numerical value, causal relation, or final conclusion. In such cases, sequence-level consistency can underestimate the true robustness failure. A robust fine-tuning objective should therefore identify and stabilise the semantic segments where perturbations induce the largest meaning shift.

Focusing on output-side semantic drift alone, however, only captures part of the picture. Transformer generation is mediated by attention mechanisms \citep{NIPS2017_3f5ee243}, and input perturbations can affect both token representations and the way contextual evidence is allocated during generation \citep{gan2024reasoning, agrawal2025enhancing}. We do not claim that attention provides a complete causal explanation of model behaviour. Instead, we use segment-level attention reallocation as a tractable diagnostic proxy: it describes how the model redistributes attention mass across source or evidence segments when the input is perturbed. This view complements the output-side semantic loss by offering a mechanism-side interpretation of why certain output segments become unstable.

In this work, we introduce Semantic Segment Robustness Regularisation (S$^2$R$^2$), a robust fine-tuning framework for LoRA-adapted LLMs \citep{hu2022lora}. S$^2$R$^2$ first decomposes clean and perturbed generations into semantic segments, aligns them through an optimal-transport objective, and penalises the segments with the largest semantic drift. It then adds an adapter-stability regulariser that controls the growth of LoRA updates in attention projections. This term should not be interpreted as a standalone new norm penalty. Rather, it serves as a tractable proxy for limiting perturbation-amplified attention reallocation under LoRA adaptation. Since LoRA updates a frozen pre-trained model through low-rank matrices, it also provides a convenient adaptation space for analysing model movement and complexity \citep{hu2022lora}.

We further connect this adapter-stability term to a PAC-Bayesian complexity view \citep{mcallester1999pac}. Under a standard interpretation where LoRA parameters define a posterior around the frozen pre-trained model, controlling adapter growth can be understood as limiting the complexity of the adapted hypothesis. We use this as a generalisation-oriented explanation rather than as a certification claim. In other words, S$^2$R$^2$ is designed to improve robustness by preserving worst-drift semantic segments while discouraging excessive adapter movement away from the pre-trained model.

\begin{figure*}[t]
\centering
\includegraphics[width=0.75\linewidth]{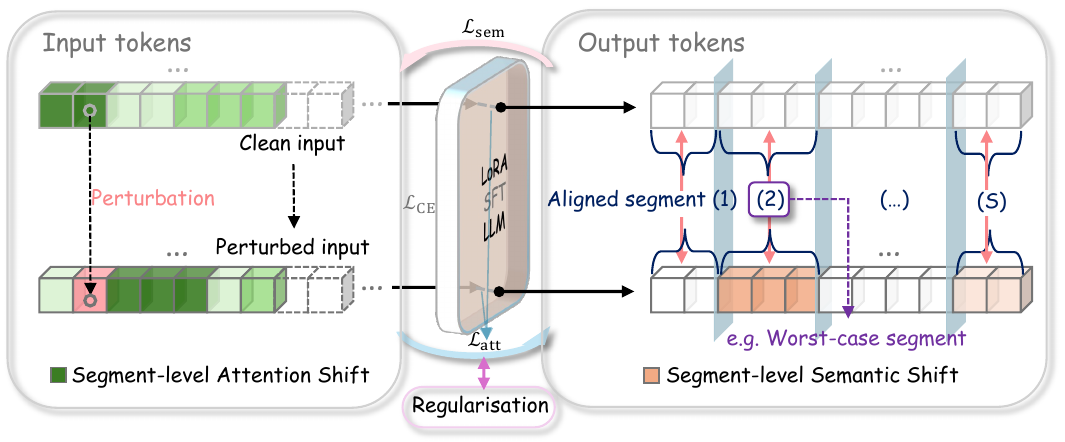}
\caption{
Overview of S$^2$R$^2$. 
The framework aligns clean and perturbed generations at the segment level and emphasises the worst-drift semantic segments. 
The mechanism-side branch, labelled as attention shift in the schematic, is implemented as an adapter-stability term $L_{\mathrm{stab}}$ that controls LoRA update growth as a proxy for limiting perturbation-amplified attention reallocation. 
Together with $L_{\mathrm{CE}}$ and $L_{\mathrm{sem}}$, this encourages robust generation by preserving critical semantic content rather than only enforcing whole-sequence consistency.
}
\label{fig:shift_visualisation}
\end{figure*}

Fig.~\ref{fig:shift_visualisation} illustrates the overall framework. A clean input and its perturbed variant are processed by a LoRA-tuned LLM. The output-side objective identifies and penalises segment-level semantic drift, while the mechanism-side stability term constrains the LoRA updates that may amplify perturbation-induced evidence reallocation. Together with the standard cross-entropy objective, these components encourage the model to preserve critical semantic content under prompt perturbations.

To summarise, the main contributions of our paper are as follows:

\textbf{C1:} We propose a segment-level view of prompt robustness, arguing that robustness failures in generation should be analysed through local semantic drift rather than only through whole-sequence consistency.

\textbf{C2:} We introduce S$^2$R$^2$, a robust LoRA fine-tuning framework that aligns clean and perturbed generations at the semantic-segment level and explicitly penalises the worst-drift segments.

\textbf{C3:} We connect segment-level robustness with adapter stability by interpreting LoRA norm control as a tractable proxy for limiting perturbation-amplified attention reallocation and as a PAC-Bayesian complexity-oriented regulariser \citep{mcallester1999pac}.\footnote{For transparency, we note that an LLM was used to assist with language polishing. See detailed statement in App.~\ref{sec:appllmuse}. The source code for this paper will be made publicly available upon acceptance.}

\textbf{C4:} We empirically evaluate S$^2$R$^2$ on summarisation benchmarks, showing improved robustness under multiple prompt perturbations and stronger cross-dataset transfer than consistency-based baselines.

\section{Preliminaries} \label{Fundamentals}

To bridge the gaps highlighted in Sec.~\ref{sec:introduction}, we first introduce three pieces of background that support our method: prompt perturbation robustness, LoRA-based fine-tuning, and the effect of perturbations on attention-based generation. Our goal in this section is not to claim a certified guarantee, but to motivate why robustness should be studied at the level of semantic segments and why LoRA provides a useful adaptation space for controlling perturbation-amplified changes.

We organise the discussion around three questions:

\textbf{Q1:} How do small prompt perturbations affect generation, and why can whole-sequence consistency hide local semantic failures?

\textbf{Q2:} Why is LoRA a suitable setting for studying robust adaptation and adapter-level stability?

\textbf{Q3:} How can perturbation-induced changes in attention be interpreted as segment-level evidence reallocation rather than as a complete causal explanation?

\subsection{Robustness to Input Perturbation}\label{relatedwork}

LLMs often exhibit sensitivity to minor, semantically preserving perturbations in the input \citep{DBLP, agrawal2025enhancing,HU2025103779}. These perturbations range from unintentional typographical errors \citep{gan2024reasoning, dongrevisit} and paraphrases \citep{wang2023largelanguagemodelsreally,hu2026lyingtruthsopenchannelmultiagent} to deliberate adversarial attacks. Although such perturbations may preserve the intended meaning from a human perspective, they can still alter the model's generated output, reduce factual consistency, or amplify unsafe behaviours.

Existing robustness methods mainly improve stability through three families of techniques. Data augmentation enriches the training set with perturbed examples so that the model is exposed to a wider range of input variations \citep{wei2019eda}, including more sophisticated augmentation strategies such as back-translation \citep{back}. Adversarial training constructs difficult examples that maximise the training loss. Since text is discrete, many methods apply projected gradient descent or related perturbations in the continuous embedding space \citep{PGD}. Consistency-based methods regularise the model by comparing its output distribution under a clean input $x$ and a perturbed input $x'$, often using KL divergence \citep{aghajanyan2021better} or Jensen-Shannon divergence \citep{qiang-etal-2024-prompt}. Representative methods such as SMART \citep{jiang-etal-2020-smart} and FreeLB \citep{Zhu2020FreeLB} further combine consistency regularisation with adversarial input construction.

These approaches have improved average robustness, but they usually treat the generated output as a holistic object. This can be limiting for generation tasks. A response may remain similar to the clean output at the sequence level while drifting on a critical entity, relation, or conclusion. Such local failures are difficult to capture with only global losses. This concern is related to findings in robust learning and adversarial fairness, where different components, groups, or classes can contribute unevenly to the total risk \citep{robustfair, jin2025enhancing}. Motivated by this non-uniformity, we study prompt robustness from the perspective of semantic segments: instead of only asking whether two full outputs are similar, we ask which parts of the generation become unstable under perturbation.

\subsection{LLMs Fine-tuning}

Parameter-Efficient Fine-Tuning (PEFT) methods reduce the cost of adapting large models by updating only a small number of task-specific parameters. Existing approaches include prompt-based methods \citep{lester-etal-2021-power} and adapter-based methods \citep{hu2022lora}. Among them, Low-Rank Adaptation (LoRA) is widely used because it keeps the pre-trained weights frozen and learns low-rank updates to selected projection matrices.

For a Transformer layer $l$, let $W^l_{0,*}$ be a frozen pre-trained projection matrix, where $* \in \{Q,K,V\}$ denotes the query, key, or value projection. LoRA parameterises the trainable update as
\[
    W^l_{*}
    =
    W^l_{0,*}
    +
    \Delta W^l_{*},
    \qquad
    \Delta W^l_{*}
    =
    B^l_{*}(A^l_{*})^\top,
\]
where $B^l_{*}\in \mathbb{R}^{d_{\mathrm{out}}\times r}$, $A^l_{*}\in \mathbb{R}^{d_{\mathrm{in}}\times r}$, and $r \ll \min(d_{\mathrm{in}}, d_{\mathrm{out}})$. Given a hidden representation $H$, the query projection becomes
\[
    Q
    =
    H
    \big(
        W^l_{0,Q}
        +
        B^l_Q(A^l_Q)^\top
    \big),
\]
and the key and value projections are updated analogously.

LoRA is suitable for our study for two reasons. First, it provides a practical setting for robust fine-tuning because only a small set of adaptation parameters is updated. Second, it gives a low-dimensional handle for analysing how fine-tuning changes attention behaviour. Since attention weights are determined by query and key projections, large LoRA updates to $Q$ and $K$ can affect how the model allocates evidence during generation. This does not mean that LoRA norm alone explains robustness, but it provides a tractable mechanism-side quantity that can be monitored and regularised.

We also use LoRA to define a complexity-oriented view of robust adaptation. Following common Bayesian treatments of neural-network parameters \citep{goodman1963statistical}, we can regard the frozen pre-trained model as defining a prior centre, and the learned LoRA parameters as a posterior movement around this centre.

\begin{assumption}[LoRA posterior with comparable variance]
The data-independent prior distribution $P=\mathcal{N}(0,\tau^2 I)$ and the data-dependent posterior distribution $Q=\mathcal{N}(\mu,\sigma^2 I)$ over LoRA adaptation parameters can be described by Gaussian distributions with comparable variances $\tau^2$ and $\sigma^2$.
\end{assumption}

This assumption is not used to claim a certified guarantee. It only supports a standard PAC-Bayesian complexity interpretation: when the adapted model moves farther from the pre-trained state, the posterior becomes more complex relative to the prior. Since LoRA restricts adaptation to a low-rank subspace, this movement can be summarised through the norms of the trainable matrices.

During LoRA fine-tuning, the two factors $A^l_*$ and $B^l_*$ are often observed to remain within a comparable scale. This behaviour is related to implicit regularisation in matrix factorisation and to empirical observations on asymmetry in low-rank adapters. We therefore use the following working assumption.

\begin{assumption}[LoRA factor-scale comparability]
For each adapted layer $l$ and projection $*$, the Frobenius norms $\|A^l_*\|_F$ and $\|B^l_*\|_F$ remain comparable in scale during fine-tuning.
\end{assumption}

This assumption is used only to interpret the product-form adapter-stability term in relation to a quadratic complexity quantity. We provide empirical validation of this comparability in the appendix.

\subsection{Perturbation Effect}

We next clarify how prompt perturbations can affect attention-based generation. For a Transformer attention head, the pre-softmax attention score matrix is
\[
    S(x)
    =
    \frac{Q(x)K(x)^\top}{\sqrt{d_k}},
\]
and the attention weight matrix is
\[
    A(x)
    =
    \mathrm{softmax}(S(x)).
\]
Here, the softmax is applied row-wise. The attention output is then obtained by
\[
    O(x)
    =
    A(x)V(x).
\]
These three objects should be distinguished: $S(x)$ is the pre-softmax score, $A(x)$ is the attention distribution, and $O(x)$ is the value-weighted context representation.

When the input representation is perturbed from $H$ to $H+\epsilon$, both the frozen pre-trained projections and the LoRA updates contribute to the resulting change in attention scores. For the query projection, we can write
\[
\begin{aligned}
    Q(x')
    &=
    (H+\epsilon)
    \big(
        W^l_{0,Q}
        +
        B^l_Q(A^l_Q)^\top
    \big)
    \\
    &=
    \underbrace{H W^l_{0,Q}}_{\text{pre-trained}}
    +
    \underbrace{H B^l_Q(A^l_Q)^\top}_{\text{task adaptation}}
    \\
    &\quad
    +
    \underbrace{\epsilon W^l_{0,Q}}_{\text{direct perturbation}}
    +
    \underbrace{\epsilon B^l_Q(A^l_Q)^\top}_{\text{adapter-mediated perturbation}} .
\end{aligned}
\]
The same decomposition applies to the key and value projections. This expression separates the perturbation effect into two conceptually different parts. The direct perturbation component comes from applying the frozen pre-trained projection to the perturbed representation. The adapter-mediated perturbation component describes how the learned LoRA update interacts with the input perturbation.

This decomposition motivates our later design. Since the model cannot remove the input perturbation itself, robust fine-tuning should avoid amplifying its effect through overly large adapter updates. At the same time, attention change should not be overstated as a full explanation of generation. Attention weights describe how the model reallocates evidence across tokens or segments; the final output also depends on value representations, decoder dynamics, and the language modelling objective. We therefore use attention reallocation as a diagnostic proxy rather than a causal explanation.

At the segment level, this proxy becomes more interpretable. Instead of comparing token-level attention matrices directly, we aggregate attention mass from output segments to source or evidence segments. This yields a segment-to-segment attention matrix that describes how evidence usage changes under perturbation. The formal definition is given in Sec.~\ref{sec:methodology}. This view connects the output-side semantic drift objective with a mechanism-side stability term: if large LoRA updates can amplify evidence reallocation, then controlling adapter growth provides a tractable way to encourage more stable robust adaptation.

\section{Methodology}
\label{sec:methodology}

\subsection{Overview}

Our goal is to make prompt-perturbation robustness both optimisable and interpretable. 
Instead of treating the generated output as a single undifferentiated sequence, we view a generation as a collection of semantic segments. 
This perspective is motivated by a simple observation: prompt perturbations rarely damage all parts of an output uniformly. 
A model may preserve most of the surface form while drifting on a key entity, relation, or conclusion. 
Therefore, a robust fine-tuning objective should not only encourage global consistency, but also identify and stabilise the output segments where the most consequential meaning drift occurs.

We instantiate this idea in the LoRA fine-tuning setting. 
LoRA provides a practical adaptation mechanism and, more importantly for our analysis, a low-dimensional handle for studying how task adaptation changes attention behaviour. 
Our framework, S$^2$R$^2$, contains two coupled components. 
First, an output-side segment alignment loss measures semantic drift between clean and perturbed generations at the segment level. 
Second, an adapter-stability regulariser controls the growth of LoRA updates that can amplify perturbation-induced attention reallocation. 
The latter should not be interpreted as a standalone new norm penalty; rather, it is used as a tractable stability proxy motivated by the segment-level attention analysis below.

\subsection{Problem Formulation}

Let $x$ denote a clean input prompt and $x'$ its perturbed version, where $x'$ may be produced by typographical noise, deletion, synonym replacement, or paraphrasing. 
Given a model $f_{\theta}$ with trainable LoRA parameters $\theta$, the model produces outputs
\[
    y = f_{\theta}(x),
    \qquad
    y' = f_{\theta}(x').
\]
The standard supervised objective minimises the cross-entropy loss on the target output. 
However, this alone does not constrain the relation between $y$ and $y'$ under perturbation. 
Consistency-based robust fine-tuning methods typically add a sequence-level penalty between the clean and perturbed predictions. 
Such holistic penalties are useful, but they can dilute local failures because all tokens or all output positions are aggregated into a single global score.

We instead decompose each output into semantic segments. 
Let
\[
    \mathcal{G}(y)=\{g_1,\ldots,g_U\},
    \qquad
    \mathcal{G}(y')=\{g'_1,\ldots,g'_{U'}\}
\]
denote the segment sets of the clean and perturbed outputs. 
In practice, segments can be obtained using lightweight punctuation-based discourse splitting, while more expensive neural segmentation can be used as a robustness check. 
Each segment is represented by the average of its token-level hidden states:
\[
    e_u
    =
    \frac{1}{|g_u|}
    \sum_{t\in g_u} h_t,
    \qquad
    e'_v
    =
    \frac{1}{|g'_v|}
    \sum_{t\in g'_v} h'_t,
\]
where $h_t$ and $h'_t$ are output-side hidden representations under the clean and perturbed inputs, respectively.

The central object of interest is therefore not only whether the whole output changes, but which semantic segments drift and how strongly they drift.

During training, we compute segment representations from teacher-forced decoder hidden states under the same reference target for the clean and perturbed inputs. Segment boundaries are treated as fixed preprocessing decisions, while gradients are back-propagated through the hidden representations used in the segment costs.

\paragraph{Segmentation granularity and neuro-symbolic semantic units.}
The formulation above is intentionally segmentation-agnostic. 
In the main experiments, we instantiate $\mathcal{G}(\cdot)$ with punctuation-based discourse splitting for efficiency, treating each segment boundary as a fixed preprocessing decision. 
However, the same objective can naturally support richer semantic units. 
A more structured alternative is to use a neuro-symbolic segmentation module that decomposes a generation into proposition-level units, where each unit corresponds to an atomic factual statement such as an entity--predicate--argument tuple, a numerical claim, a causal relation, a contrastive relation, or a conclusion-bearing statement. 
Under this view, a segment is not merely a span separated by punctuation, but a semantically meaningful proposition whose drift is likely to affect the factual content of the whole response.

This extension can be incorporated without changing the differentiable training objective. 
The neuro-symbolic module only determines segment spans and optional importance weights before optimisation, while gradients are still back-propagated through the teacher-forced hidden representations used to compute the segment costs. 
For a segment $g_u$, one may define an importance weight
\[
\begin{aligned}
    w_u
    =
    \mathrm{Norm}\big(
    &\alpha_{\mathrm{ent}} I_u^{\mathrm{ent}}
    +
    \alpha_{\mathrm{num}} I_u^{\mathrm{num}}
    \\
    &+
    \alpha_{\mathrm{rel}} I_u^{\mathrm{rel}}
    +
    \alpha_{\mathrm{cen}} I_u^{\mathrm{cen}}
    \big).
\end{aligned}
\]
where $I_u^{\mathrm{ent}}$, $I_u^{\mathrm{num}}$, $I_u^{\mathrm{rel}}$, and $I_u^{\mathrm{cen}}$ indicate whether the segment contains named entities, numerical quantities, relational predicates, or high summary-level centrality, respectively. 
The transport marginals in Sec.~\ref{sec:methodology} can then be replaced by these importance weights, yielding an importance-aware alignment objective. 
This makes it possible to assign larger robustness weight to factual or conclusion-bearing propositions than to peripheral segments. 
We leave this neuro-symbolic instantiation to future experiments and use the lightweight segmentation strategy in the present empirical study.

\textbf{A detailed rule set for the neuro-symbolic segmentation module, including proposition extraction, importance weighting, and symbolic mismatch scoring, will be provided in an updated preprint version.}

\subsection{Segment-Level Semantic Drift}

To compare the clean and perturbed segment sets, a fixed one-to-one alignment is brittle because perturbations may change the order, length, or granularity of generated text. 
We therefore align the two segment sets using an optimal transport plan. 
Let the semantic cost between segment $g_u$ and $g'_v$ be
\[
    c_{uv}
    =
    1-\cos(e_u,e'_v).
\]
We compute an alignment matrix
\[
    T^{\star}
    \in
    [0,1]^{U\times U'}
\]
by solving the transport problem
\[
    T^{\star}
    =
    \arg\min_{T\in\Pi(\mathcal{G}(y),\mathcal{G}(y'))}
    \sum_{u=1}^{U}
    \sum_{v=1}^{U'}
    T_{uv} c_{uv},
\]
where $\Pi(\mathcal{G}(y),\mathcal{G}(y'))$ denotes the set of admissible transport plans between the two empirical segment distributions.

The segment-level drift for a clean segment $g_u$ is then
\[
    d_u
    =
    \sum_{v=1}^{U'} T^{\star}_{uv} c_{uv}.
\]
Instead of only averaging these drift values, we place additional emphasis on the most unstable segments. 
The semantic robustness loss is defined as
\[
    \mathcal{L}_{\mathrm{sem}}
    =
    \frac{1}{U}
    \sum_{u=1}^{U} d_u
    +
    \frac{1}{\beta}
    \log
    \sum_{u=1}^{U}
    \exp(\beta d_u),
\]
where $\beta>0$ controls the strength of the smooth maximum. 
The first term encourages overall segment-level consistency, while the second term focuses optimisation on the worst-drift segments. 
This design directly reflects the intuition that a small number of semantically important segments can dominate the robustness failure of the whole generation.

\subsection{Segment-Level Attention Reallocation}

The semantic loss above captures output-side drift. 
To understand how perturbations may lead to such drift inside the model, we analyse attention reallocation at the segment level. 
For a transformer layer $\ell$ and attention head $h$, let
\[
    A^{\ell,h}(x)
    \in
    \mathbb{R}^{T_q\times T_k}
\]
denote the attention matrix under the clean input, where $T_q$ is the number of query positions and $T_k$ is the number of key positions. 
The corresponding matrix under the perturbed input is $A^{\ell,h}(x')$.

We aggregate token-level attention into a segment-to-segment matrix. 
Let $\{g_1,\ldots,g_U\}$ be output segments and $\{s_1,\ldots,s_V\}$ be source or evidence segments. 
For encoder-decoder models, the query segments are decoder-side output segments and the key segments are encoder-side source segments. 
For decoder-only models, the same definition applies over the causally available input and prefix tokens. 
We define
\[
    C^{\ell,h}_{uv}(x)
    =
    \frac{1}{|g_u|}
    \sum_{i\in g_u}
    \sum_{j\in s_v}
    A^{\ell,h}_{ij}(x).
\]
Thus, $C^{\ell,h}_{uv}(x)$ measures how much attention mass the model assigns to source segment $s_v$ when producing output segment $g_u$. 
Each row $C^{\ell,h}_{u,:}(x)$ can be interpreted as a coarse-grained distribution over evidence segments.

The perturbation-induced attention reallocation for output segment $g_u$ is measured by
\[
    r^{\ell,h}_u
    =
    \mathrm{JS}
    \big(
        C^{\ell,h}_{u,:}(x)
        \,\Vert\,
        C^{\ell,h}_{u,:}(x')
    \big),
\]
where $\mathrm{JS}(\cdot\Vert\cdot)$ is the Jensen-Shannon divergence. 
A large value of $r^{\ell,h}_u$ indicates that, when generating the same semantic part of the output, the model relies on a substantially different distribution of input evidence after perturbation.

This statistic is not intended to claim that attention is a complete causal explanation of generation. 
Rather, it provides a tractable diagnostic proxy for how perturbations redistribute evidence usage across semantic segments. 
To clarify the role of this proxy, the context representation of an output segment can be approximated as
\[
    z_u(x)
    \approx
    \sum_{v=1}^{V}
    C_{uv}(x)\,\bar{v}_v(x),
\]
where $\bar{v}_v(x)$ is the average value representation of source segment $s_v$. 
Therefore, the segment representation shift can be decomposed as
\[
\begin{aligned}
    z_u(x') - z_u(x)
    \approx
    &
    \underbrace{
    \sum_{v=1}^{V}
    \big(
        C_{uv}(x') - C_{uv}(x)
    \big)
    \bar{v}_v(x)
    }_{\text{attention reallocation effect}}
    \\
    &
    +
    \underbrace{
    \sum_{v=1}^{V}
    C_{uv}(x')
    \big(
        \bar{v}_v(x')-\bar{v}_v(x)
    \big)
    }_{\text{value representation drift}}.
\end{aligned}
\]
This decomposition shows that attention reallocation is only one source of semantic drift, but it is a useful mechanism-side signal that complements the output-side segment loss.

\subsection{LoRA Adapter Stability as a Tractable Proxy}

Directly regularising all attention matrices across layers, heads, and perturbation types is computationally expensive and may introduce unstable gradients. 
Instead, we control a tractable source of attention reallocation: the magnitude of the LoRA updates to the attention projections.

For a transformer layer $\ell$, LoRA updates a projection matrix $W_*^\ell$ by
\[
    W_*^\ell
    =
    W_{0,*}^\ell
    +
    \Delta W_*^\ell,
    \qquad
    \Delta W_*^\ell
    =
    B_*^\ell (A_*^\ell)^{\top},
\]
where $*\in\{Q,K,V\}$ denotes query, key, or value projections. 
The pre-trained matrix $W_{0,*}^\ell$ remains frozen, while $A_*^\ell$ and $B_*^\ell$ are trainable low-rank matrices.

For attention weights, the query and key projections are the most direct source of pre-softmax score changes:
\[
    S^{\ell,h}(x)
    =
    \frac{Q^{\ell,h}(x)K^{\ell,h}(x)^{\top}}{\sqrt{d_k}}.
\]
Perturbations to the input and changes in the LoRA matrices both affect $S^{\ell,h}(x)$, which is then transformed by the softmax function into attention weights. 
Using the local Lipschitz property of softmax and standard matrix norm inequalities, the resulting segment-level attention reallocation can be upper-bounded by terms depending on the input perturbation magnitude and the norms of the LoRA updates. 
Informally,
\[
\begin{aligned}
    \|C^{\ell,h}(x')-C^{\ell,h}(x)\|_F
    &\leq
    \Gamma_{\epsilon}^{\ell,h}
    \\
    \quad+
    \Gamma_{\mathrm{LoRA}}^{\ell,h}
    \Big(
        \|\Delta W_Q^\ell\|_F
        +
        \|\Delta W_K^\ell\|_F
    \Big).
\end{aligned}
\]
where $\Gamma_{\epsilon}^{\ell,h}$ captures the uncontrollable contribution of the input perturbation, and $\Gamma_{\mathrm{LoRA}}^{\ell,h}$ collects layer-dependent constants. 
Since
\[
    \|\Delta W_*^\ell\|_F
    =
    \|B_*^\ell (A_*^\ell)^{\top}\|_F
    \leq
    \|B_*^\ell\|_F
    \|A_*^\ell\|_F,
\]
we introduce the adapter-stability regulariser
\[
    \mathcal{L}_{\mathrm{stab}}
    =
    \sum_{\ell}
    \sum_{*\in\{Q,K\}}
    \|B_*^\ell\|_F
    \|A_*^\ell\|_F.
\]
When value-state stability is also desired, the same term can be extended to include $V$. 
In our main formulation, we focus on $Q$ and $K$ because they directly determine attention reallocation.

This term should be understood as a stability proxy rather than as a direct attention loss. 
It does not replace the segment-level semantic objective; instead, it discourages the adapted model from relying on large low-rank updates that can amplify perturbation-induced evidence reallocation.

\subsection{Complexity Interpretation}

The adapter-stability term also admits a natural complexity interpretation. 
Consider a PAC-Bayesian view in which the LoRA parameters define a posterior distribution $Q$ over adapted models, while a data-independent prior $P$ is centred around the pre-trained model with zero LoRA update. 
For an isotropic Gaussian prior and posterior with fixed or comparable variances, the KL term can be written up to constants as
\[
    \mathrm{KL}(Q\Vert P)
    \approx
    \frac{1}{2\tau^2}
    \sum_{\ell}
    \sum_{*}
    \Big(
        \|B_*^\ell\|_F^2
        +
        \|A_*^\ell\|_F^2
    \Big).
\]
The product-form stability term is connected to this complexity quantity through
\[
\begin{aligned}
     \|B_*^\ell\|_F^2
    +
    \|A_*^\ell\|_F^2
    &=
    \big(
        \|B_*^\ell\|_F
        -
        \|A_*^\ell\|_F
    \big)^2 \\
    &+
    2
    \|B_*^\ell\|_F
    \|A_*^\ell\|_F.
\end{aligned}
\]
Thus, when the two LoRA factors remain balanced in magnitude, controlling
$\|B_*^\ell\|_F\|A_*^\ell\|_F$
also controls the dominant part of the LoRA complexity. 
If stronger control is required, one can add an explicit imbalance penalty
\[
    \mathcal{L}_{\mathrm{bal}}
    =
    \sum_{\ell}
    \sum_{*}
    \big(
        \|B_*^\ell\|_F
        -
        \|A_*^\ell\|_F
    \big)^2.
\]
In the main experiments, we monitor the LoRA norms to verify that the learned factors remain comparable.

\begin{figure}[t]
\centering
\includegraphics[width=\linewidth]{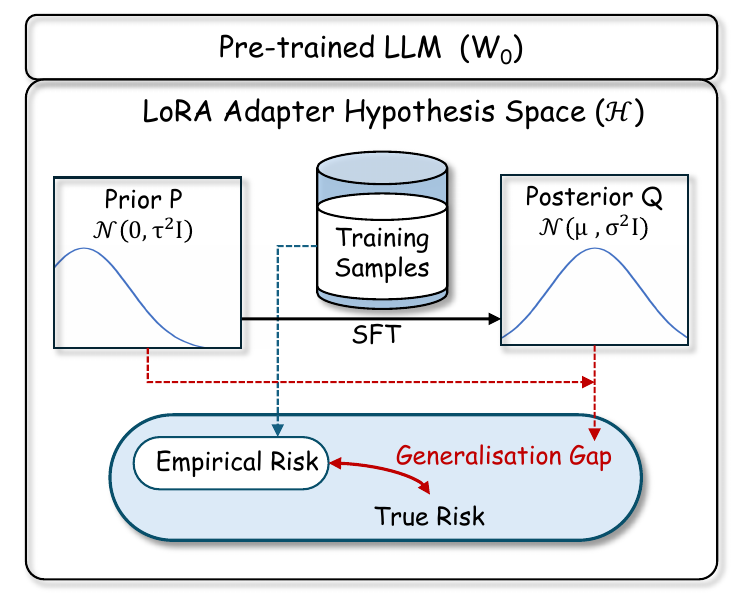}
\caption{
PAC-Bayesian complexity view of LoRA adaptation. 
The frozen pre-trained model defines the prior centre, while LoRA fine-tuning induces a posterior over a restricted adapter hypothesis space. 
We use this view to interpret the adapter-stability term as a complexity-oriented control on the movement from the pre-trained model. 
The figure is intended as an explanatory view of complexity-adjusted robustness, not as a standalone certification claim.
}
\label{fig:lora_pac}
\end{figure}

We use this PAC-Bayesian view as a generalisation-oriented interpretation, not as a standalone certification claim. 
It explains why a robustness objective that avoids excessive adapter growth may transfer better to unseen perturbations or unseen datasets: the model is encouraged to improve segment-level stability without moving too far from the pre-trained hypothesis space.
Fig.~\ref{fig:lora_pac} illustrates this interpretation. The prior corresponds to the frozen pre-trained model with zero LoRA update, while the posterior corresponds to the adapted model after fine-tuning. The empirical risk captures observed robustness under perturbations, and the complexity term reflects the cost of moving within the LoRA adapter hypothesis space.

\subsection{Training Objective}

The final S$^2$R$^2$ objective combines the standard task loss, segment-level semantic robustness, and adapter stability:
\[
    \mathcal{L}_{\mathrm{S^2R^2}}
    =
    \mathcal{L}_{\mathrm{CE}}
    +
    \lambda_{\mathrm{sem}}
    \mathcal{L}_{\mathrm{sem}}
    +
    \lambda_{\mathrm{stab}}
    \mathcal{L}_{\mathrm{stab}}.
\]
When the optional balance term is used, the objective becomes
\[
    \mathcal{L}_{\mathrm{S^2R^2}}
    =
    \mathcal{L}_{\mathrm{CE}}
    +
    \lambda_{\mathrm{sem}}
    \mathcal{L}_{\mathrm{sem}}
    +
    \lambda_{\mathrm{stab}}
    \mathcal{L}_{\mathrm{stab}}
    +
    \lambda_{\mathrm{bal}}
    \mathcal{L}_{\mathrm{bal}}.
\]

This objective has a clear division of labour. 
The cross-entropy term preserves task performance. 
The semantic segment loss directly reduces output-side meaning drift under perturbations. 
The adapter-stability term controls a mechanism-side source of attention reallocation and provides a complexity-aware regularisation effect. 
Together, these terms encourage the model to produce robust outputs by preserving critical semantic segments rather than merely enforcing global sequence similarity.

\section{Experiment}
\label{sec:experiment}

\subsection{Experimental Setup}

\paragraph{Task selection.}
We use summarisation as the main evaluation task because it provides long-form generations with rich semantic structure. Compared with shorter classification-style outputs, summarisation makes it easier to observe whether perturbations affect the whole response uniformly or concentrate their effect on specific semantic segments. This setting is therefore well aligned with our segment-level view of prompt robustness.

\paragraph{Models and datasets.}
We evaluate S$^2$R$^2$ on both encoder-decoder and decoder-only architectures, including BART-base, Flan-T5-base, and Mistral-7B-Instruct-v0.2. The experiments cover three summarisation benchmarks with different domain and style properties: CNN/DailyMail, XSum, and PubMed. CNN/DailyMail is relatively extractive and useful for testing factual stability; XSum is more abstractive and stresses semantic coherence; PubMed contains technical biomedical content and tests domain-specific precision.

\paragraph{Baselines and implementation.}
We compare against R3F and SMART, two representative robustness fine-tuning baselines discussed in Sec.~\ref{relatedwork}. All methods are fine-tuned using LoRA to ensure a controlled comparison under the same parameter-efficient adaptation setting. The base training objective is the standard cross-entropy loss $\mathcal{L}_{\mathrm{CE}}$. For S$^2$R$^2$, we add the semantic segment loss $\mathcal{L}_{\mathrm{sem}}$ to directly reduce clean--perturbed segment drift. We also include the adapter-stability term $\mathcal{L}_{\mathrm{stab}}$, which controls the growth of LoRA updates and serves as a complexity-oriented diagnostic under the PAC-Bayesian interpretation. The parameter update procedure is summarised in Algorithm~\ref{alg:s2r2} in the appendix.

For the main experiments, we use a computationally efficient punctuation-based segmentation strategy. To check whether this lightweight segmentation introduces a substantial bias, we also conduct a smaller-scale validation with a T5-based semantic segmentation method. The results in the appendix show similar convergence behaviour, while the punctuation-based method is substantially more efficient.

\subsection{Evaluation}

\subsubsection{Perturbation Testbed}

To simulate common real-world input corruptions, we apply three types of perturbations to the source articles in the test sets, following prior work on prompt perturbation robustness \citep{qiang-etal-2024-prompt,dongrevisit,wang2023largelanguagemodelsreally}.

\paragraph{Typographical noise and deletion.}
We introduce character-level typographical noise by randomly swapping characters within words, and then delete a subset of words from the source text. This branch tests whether the model can remain stable under noisy or partially corrupted inputs.

\paragraph{Synonym replacement.}
We replace selected words with synonyms to evaluate whether the model preserves output semantics under local lexical variation. This perturbation is designed to alter surface form while largely preserving the intended meaning.

\paragraph{Paraphrasing.}
We use a pre-trained paraphrasing model to rewrite source texts. This perturbation is more challenging because it can substantially change syntax and lexical choices while maintaining the original semantic content.

\subsubsection{Evaluation Metrics}

We evaluate both task performance and clean--perturbed output consistency.

\paragraph{Performance Drop Rate.}
Performance Drop Rate (PDR) measures the relative degradation in ROUGE-L under perturbation:
\[
    \mathrm{PDR}_{p}
    =
    1 -
    \frac{
        R_L(f_p)
    }{
        R_L(f_c)
    },
\]
where $R_L(f_c)$ and $R_L(f_p)$ are the ROUGE-L scores of the clean and perturbed outputs, respectively. Values closer to zero indicate smaller performance degradation.

\paragraph{Self-BERTScore.}
Self-BERTScore (SB) measures semantic consistency between the clean output $f_c$ and perturbed output $f_p$ using BERTScore \citep{2020BERTScore:}. We report $1-\mathrm{SB}$ so that lower values indicate stronger semantic stability.

\paragraph{Output edit rate.}
Output edit rate $\Delta_{\mathrm{ed}}$ measures surface-form instability between $f_c$ and $f_p$ using the normalised word-level Levenshtein distance \citep{Levenshteindistance}. Lower values indicate smaller syntactic change in the generated output.

\paragraph{Empirical risk and complexity-adjusted score.}
To summarise robustness across the above metrics, we define a heuristic empirical risk:
\[
    \mathrm{E\text{-}Risk}
    =
    0.8(1-\mathrm{SB})
    +
    0.1\lvert \mathrm{PDR} \rvert
    +
    0.1\Delta_{\mathrm{ed}}.
\]
We assign a larger weight to semantic stability because S$^2$R$^2$ is designed to preserve meaning under perturbation. Other normalised combinations are possible; this score is used only as an aggregate empirical indicator.

We also report $D_{\mathrm{KL}}$ and PAC-B in the tables. In the revised interpretation of this paper, these values should be read as \emph{complexity-oriented diagnostics} rather than standalone certification claims. They indicate whether robustness is achieved with limited LoRA adapter movement. A lower value suggests that the model improves robustness without relying on excessive perturbation-specific parameter updates.

\subsection{Results and Analysis}

\begin{table*}[!t]
    \centering
    \caption{Main experimental results. $|\text{PDR}|_\text{avg}$, $\Delta_{ed\text{-avg}}$, and 1-SB$_\text{avg}$ are averaged over all perturbation types. E-Risk is the empirical risk, and PAC-B denotes a PAC-Bayesian complexity-adjusted diagnostic computed from the empirical risk and the LoRA-norm-based complexity term. We report it as an indicator of robustness obtained with limited adapter movement. The best result in each category is in \textbf{bold}. Lower is better for all metrics.}
    \label{tab:detail}
    \renewcommand{\arraystretch}{1.2} 
    
    \resizebox{\textwidth}{!}{
    \begin{tabular}{lccc>{\columncolor{gray!7}}c ccc>{\columncolor{gray!7}}c ccc>{\columncolor{gray!7}}c ccc}
    \toprule
    \multirow{2}{*}{Method} &
    \multicolumn{4}{c}{$|\text{PDR}| \downarrow$} &
    \multicolumn{4}{c}{1-SB $\downarrow$} &
    \multicolumn{4}{c}{$\Delta_{ed} \downarrow$} &
    \multirow{2}{*}{$D_\text{KL}$ $\downarrow$} &
    \multirow{2}{*}{E-Risk $\downarrow$} &
    \multirow{2}{*}{PAC-B $\downarrow$} \\
    \cmidrule(lr){2-5}\cmidrule(lr){6-9}\cmidrule(lr){10-13}
      & Typo & Syno & Para & Avg & Typo & Syno & Para & Avg & Typo & Syno & Para & Avg &  &  &  \\
    \midrule

    \multicolumn{16}{c}{\textbf{(a) Bart-base on CNN/Dailymail}} \\
    \cmidrule(lr){1-16}
    R3F                 & 0.0949 & 0.0613 & 0.0872 & 0.0811 & 0.6402 & 0.4795 & 0.7507 & 0.6238 & 0.7524 & 0.6653 & 0.8597 & 0.7591 & 233.625 & 0.5831 & 0.6930 \\
    SMART               & 0.0008 & 0.0011 & 0.0176 & 0.0064 & 0.0611 & 0.0663 & 0.4176 & \textbf{0.1817} & 0.1378 & 0.1434 & 0.8806 & \textbf{0.3873} & 202.554 & 0.1821 & 0.2874 \\
    \rowcolor{gray!10}
    \textbf{S$^2$R$^2$} & 0.0006 & 0.0015 & 0.0057 & \textbf{0.0012} & 0.0717 & 0.0661 & 0.4470 & 0.1956 & 0.1674 & 0.1548 & 0.8892 & 0.4038 & \textbf{78.006} & 0.1966 & \textbf{0.2626} \\
    \midrule

    \multicolumn{16}{c}{\textbf{(b) Bart-base on XSum}} \\
    \cmidrule(lr){1-16}
    R3F                 & 0.0320 & 0.0188 & 0.0548 & 0.0352 & 0.4166 & 0.2905 & 0.5898 & 0.4323 & 0.6748 & 0.5091 & 0.8985 & 0.6914 & 190.051 & 0.4185 & 0.5181 \\
    SMART               & 0.0108 & 0.0004 & 0.0264 & \textbf{0.0125} & 0.0566 & 0.1076 & 0.4014 & 0.2220 & 0.3196 & 0.2292 & 0.7558 & \textbf{0.4349} & 149.155 & 0.2223 & 0.3110 \\
    \rowcolor{gray!11}
    \textbf{S$^2$R$^2$} & 0.0346 & 0.0138 & 0.0072 & 0.0185 & 0.0506 & 0.1144 & 0.3766 & \textbf{0.2141} & 0.3912 & 0.2962 & 0.7775 & 0.4883 & \textbf{90.775} & \textbf{0.2219} & \textbf{0.2924} \\
    \midrule

    \multicolumn{16}{c}{\textbf{(c) T5-base on PubMed}} \\
    \cmidrule(lr){1-16}
    R3F                 & 0.1266 & 0.0327 & 0.1043 & 0.0781 & 0.6613 & 0.3310 & 0.7105 & 0.5676 & 0.7225 & 0.3743 & 0.7716 & 0.6228 & 1377.258 & 0.5242 & 0.7874 \\
    SMART               & 0.0509 & 0.0060 & 0.0515 & \textbf{0.0321} & 0.6435 & 0.3737 & 0.8035 & 0.6089 & 0.6812 & 0.3944 & 0.8628 & 0.6461 & 1000.989 & 0.5560 & 0.7795 \\
    \rowcolor{gray!11}
    \textbf{S$^2$R$^2$} & 0.3229 & 0.1592 & -- & -- & 0.1281 & 0.1174 & 0.2873 & \textbf{0.1776} & 0.0766 & 0.0702 & 0.2116 & \textbf{0.1195} & \textbf{774.055} & \textbf{0.2355} & \textbf{0.4333} \\
    \midrule

    \multicolumn{16}{c}{\textbf{(d) Mistral-7B on PubMed}} \\
    \cmidrule(lr){1-16}
    R3F                 & 0.0893 & 0.0568 & 0.2421 & 0.1300 & 0.5011 & 0.2144 & 0.8413 & 0.5189 & 0.2454 & 0.1419 & 0.7842 & \textbf{0.3902} & 565.093 & 0.4672 & 0.6365 \\
    SMART               & 0.0894 & 0.0588 & 0.2221 & 0.1235 & 0.5208 & 0.2119 & 0.7317 & \textbf{0.4881} & 0.2426 & 0.1376 & 0.7905 & 0.3903 & 461.931 & 0.4419 & 0.5951 \\
    \rowcolor{gray!11}
    \textbf{S$^2$R$^2$} & 0.0795 & 0.0553 & 0.0902 & \textbf{0.0749} & 0.5499 & 0.2674 & 0.8417 & 0.5530 & 0.3444 & 0.2472 & 0.7734 & 0.4550 & \textbf{58.106} & 0.4419 & \textbf{0.5530} \\
    \bottomrule
    \end{tabular}}

\vspace{10pt}
    \caption{Zero-shot experimental result. Cross-dataset evaluation to assess the transferability of learned robustness. Bart-base are fine-tuned and evaluated on (a) different domains (general news to biomedical) and (b) different task styles (abstractive to extractive summarisation). }
    \label{tab:zero-shot}
    
    \renewcommand{\arraystretch}{1.2} 
    
    \resizebox{\textwidth}{!}{
    \begin{tabular}{lccc>{\columncolor{gray!7}}c ccc>{\columncolor{gray!7}}c ccc>{\columncolor{gray!7}}c ccc}
    \toprule
    \multirow{2}{*}{Method} &
    \multicolumn{4}{c}{$|\text{PDR}|$ $\downarrow$} &
    \multicolumn{4}{c}{$\Delta_{ed}$ $\downarrow$} &
    \multicolumn{4}{c}{1-SB $\downarrow$} &
    \multirow{2}{*}{$D_\text{KL}$ $\downarrow$} &
    \multirow{2}{*}{E-Risk $\downarrow$} &
    \multirow{2}{*}{PAC-B $\downarrow$} \\
    \cmidrule(lr){2-5}\cmidrule(lr){6-9}\cmidrule(lr){10-13}
      & Typo & Syno & Para & Avg & Typo & Syno & Para & Avg & Typo & Syno & Para & Avg &  &  & \\
    \midrule

    \multicolumn{16}{c}{\textbf{(a) Fine-tuned on CNN/DailyMail tested on PubMed}} \\
    \cmidrule(lr){1-16}
    R3F                 & 0.0952 & 0.0496 & 0.1164 & 0.0871 & 0.6649 & 0.4487 & 0.7442 & 0.6497 & 0.6461 & 0.5148 & 0.7881 & 0.6292 & 233.625 & 0.5770 & 0.6870 \\
    SMART               & 0.0138 & 0.0032 & 0.1392 & 0.0408 & 0.1921 & 0.3943 & 0.6124 & 0.6246 & 0.4204 & 0.4198 & 1.0340 & 0.3329 & 202.554 & 0.3328 & 0.4355 \\
    \rowcolor{gray!11}
    \textbf{S$^2$R$^2$} & 0.0012 & 0.0001 & 0.0662 & \textbf{0.0217} & 0.1014 & 0.0918 & 0.3665 & \textbf{0.4242} & 0.2594 & 0.2361 & 0.7771 & \textbf{0.1866} & \textbf{78.006} & \textbf{0.1939} & \textbf{0.2595} \\
    \midrule

    \multicolumn{16}{c}{\textbf{(b) Fine-tuned on XSum tested on CNN/DailyMail}} \\
    \cmidrule(lr){1-16}
    R3F                 & 0.0453 & 0.0357 & 0.3634 & 0.0941 & 0.6197 & 0.6023 & 0.7846 & 0.8321 & 0.8195 & 0.8052 & 0.8716 & 0.6689 & 190.051 & 0.6277 & 0.7273 \\
    SMART               & 0.0031 & 0.0004 & 0.0366 & 0.0113 & 0.1264 & 0.1109 & 0.8114 & 0.5839 & 0.3193 & 0.2861 & 1.1460 & 0.3496 & 149.155 & 0.3392 & 0.4279 \\
    \rowcolor{gray!11}
    \textbf{S$^2$R$^2$} & 0.0003 & 0.0027 & 0.0132 & \textbf{0.0052} & 0.1159 & 0.1004 & 0.5995 & \textbf{0.5152} & 0.3049 & 0.2636 & 0.9771 & \textbf{0.2719} & \textbf{90.775} & \textbf{0.2695} & \textbf{0.3399} \\
    \bottomrule
    \end{tabular}}

\end{table*}

\subsubsection{Robustness on Standard Benchmarks}

Table~\ref{tab:detail} reports the main results on CNN/DailyMail and XSum. 
On CNN/DailyMail with BART-base, S$^2$R$^2$ achieves a very small average PDR of 0.0012, corresponding to an 81\% reduction compared with SMART. 
This indicates that the proposed segment-level objective substantially reduces ROUGE-L degradation under perturbations. 
Although the average semantic-consistency score is close to SMART, S$^2$R$^2$ obtains this robustness with a much smaller LoRA complexity term, reducing $D_{\mathrm{KL}}$ from 202.554 to 78.006. 
This suggests that S$^2$R$^2$ does not rely on large adapter movement to fit the perturbed examples, but instead encourages a more conservative adaptation trajectory.

On XSum, which is more abstractive and therefore more sensitive to local semantic changes, S$^2$R$^2$ again obtains the lowest complexity-adjusted PAC-B value among the compared methods. 
The average PDR is slightly higher than SMART, but S$^2$R$^2$ achieves comparable semantic stability while substantially reducing the LoRA complexity term. 
This trade-off is consistent with our revised interpretation: the method is not simply optimising one surface metric, but preserving segment-level semantic stability while avoiding excessive perturbation-specific parameter updates.

Overall, the standard benchmark results show that S$^2$R$^2$ improves or preserves robustness across multiple perturbation types while maintaining a smaller adapter-complexity footprint. 
This supports the central claim that segment-level semantic alignment can provide a more targeted robustness signal than purely holistic consistency objectives.

\subsubsection{Robustness in Specialised Domains}

The PubMed experiments provide a more challenging domain-specific stress test. 
Biomedical summaries contain specialised entities, technical terminology, and fine-grained factual relations, so surface-level perturbations can interact with the generation process in more complex ways. 
With T5-base, S$^2$R$^2$ exhibits a high ROUGE-based PDR under paraphrasing. 
However, this is paired with much lower $\Delta_{\mathrm{ed}}$ and $1-\mathrm{SB}$ values than the baselines. 
This indicates that the generated outputs remain semantically close to their clean-input counterparts even when the ROUGE score drops. 
In this setting, the ROUGE degradation should therefore be interpreted carefully: it may partly reflect valid paraphrastic variation rather than purely semantic failure.

The Mistral-7B-Instruct results show a complementary pattern. 
S$^2$R$^2$ achieves empirical risk comparable to the strongest baseline while reducing $D_{\mathrm{KL}}$ by a large margin. 
This suggests that R3F and SMART may obtain robustness through larger adapter movement, whereas S$^2$R$^2$ encourages a more stable adaptation path. 
The lower complexity-adjusted score therefore supports the adapter-stability interpretation: the method improves robustness without moving too far from the pre-trained model.

Taken together, the specialised-domain results suggest that S$^2$R$^2$ is especially useful when robustness cannot be judged by n-gram overlap alone. 
The segment-level objective helps preserve semantic content under perturbation, while the adapter-stability term discourages overfitting to the perturbation patterns seen during fine-tuning.

\subsubsection{Visualisation of Adapter Stability}

\begin{figure}[t]
\centering
\includegraphics[width=\linewidth]{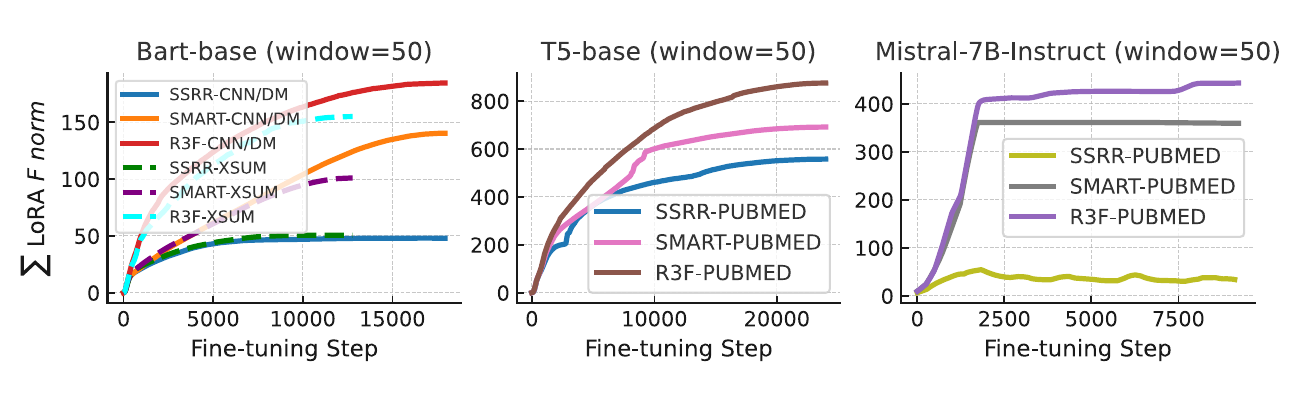}
\captionsetup{width=\linewidth}
\caption{Evolution of the sum of LoRA Frobenius-norm products during fine-tuning. S$^2$R$^2$ maintains lower adapter growth than the baselines across models and datasets. This supports the adapter-stability interpretation: robustness is improved without relying on excessive LoRA parameter movement. The x-axis differs across subfigures because the datasets and training batches have different sizes.}
\label{fig:norm_evolution}
\end{figure}

Fig.~\ref{fig:norm_evolution} visualises the adapter-stability behaviour during fine-tuning. Across the evaluated settings, S$^2$R$^2$ keeps the LoRA norm product
$\sum_l \|B^l\|_F \|A^l\|_F$
lower than the baselines. This observation is consistent with the complexity interpretation in Sec.~\ref{sec:methodology}: the segment-level objective improves robustness while the stability term discourages excessive movement away from the pre-trained model. Importantly, this figure should not be read as proof that norm control alone explains robustness. Rather, it supports the claim that S$^2$R$^2$ combines output-side semantic alignment with more conservative adapter adaptation.

\subsubsection{Zero-Shot Cross-Dataset Generalisation}

To test whether the learned robustness transfers beyond the source dataset, we conduct zero-shot cross-dataset evaluation. The model is fine-tuned on one summarisation dataset and then directly evaluated under perturbations on another unseen dataset.

When fine-tuned on CNN/DailyMail and tested on PubMed, S$^2$R$^2$ achieves the best aggregate empirical metrics, including the lowest E-Risk among the compared methods. This result suggests that the segment-level objective learns a robustness pattern that transfers from general news summarisation to a specialised biomedical domain. When fine-tuned on XSum and tested on CNN/DailyMail, S$^2$R$^2$ again outperforms the baselines across the main empirical metrics. This second setting tests transfer across summarisation styles, from highly abstractive to more extractive generation.

Overall, the zero-shot results provide empirical evidence that preserving worst-drift semantic segments can support robustness transfer beyond the perturbations and datasets observed during fine-tuning. The accompanying lower complexity scores further suggest that this transfer is achieved without large perturbation-specific LoRA updates.

\section{Conclusion}

This paper revisits prompt perturbation robustness from a segment-level perspective. Instead of treating a generated response as a single holistic object, we argue that robustness failures often concentrate on a small number of semantically critical segments. We propose S$^2$R$^2$, a robust LoRA fine-tuning framework that aligns clean and perturbed generations at the semantic-segment level and emphasises the worst-drift segments. To connect this output-side objective with model adaptation, we introduce an adapter-stability term motivated by segment-level attention reallocation and interpret it through a PAC-Bayesian complexity view.

Experiments on summarisation benchmarks show that S$^2$R$^2$ improves robustness under typographical noise, deletion, synonym replacement, and paraphrasing while maintaining competitive clean performance. The method also shows stronger cross-dataset transfer than consistency-based baselines, suggesting that segment-level semantic preservation is a useful principle for robust language-model adaptation. Future work will extend this framework with direct attention-reallocation diagnostics and additional ablations, including norm-only and segment-only variants, to further separate the contribution of semantic alignment from adapter stability.

\bibliography{anthology}

\appendix
\section{Additional Discussion on Adapter Stability}
\label{app:adapter_discussion}

This appendix provides additional intuition for the adapter-stability term used in the main text. 
The goal is not to establish a certified robustness guarantee, but to clarify why controlling LoRA growth is a reasonable mechanism-side stability proxy.

For a transformer attention head, the pre-softmax attention score is
\[
    S(x)
    =
    \frac{Q(x)K(x)^\top}{\sqrt{d_k}},
\]
where $Q(x)$ and $K(x)$ are the query and key projections. 
Under LoRA adaptation, a projection matrix is written as
\[
    W_*
    =
    W_{0,*}
    +
    \Delta W_*,
    \qquad
    \Delta W_*
    =
    B_*A_*^\top,
\]
where $W_{0,*}$ is frozen and $A_*,B_*$ are trainable low-rank matrices.

When the input representation changes from $H$ to $H+\epsilon$, the query projection can be decomposed as
\[
\begin{aligned}
    Q(x')
    &=
    (H+\epsilon)
    \big(
        W_{0,Q}
        +
        B_QA_Q^\top
    \big)
    \\
    &=
    HW_{0,Q}
    +
    HB_QA_Q^\top
    +
    \epsilon W_{0,Q}
    +
    \epsilon B_QA_Q^\top .
\end{aligned}
\]
The last term shows that the learned adapter can interact with the input perturbation. 
If the LoRA update becomes large, this adapter-mediated component can amplify the effect of perturbations on attention scores. 
This motivates controlling the magnitude of $\Delta W_Q$ and $\Delta W_K$.

Using the submultiplicative property of the Frobenius norm,
\[
    \|\Delta W_*\|_F
    =
    \|B_*A_*^\top\|_F
    \leq
    \|B_*\|_F\|A_*\|_F .
\]
Thus, the product $\|B_*\|_F\|A_*\|_F$ provides a tractable upper-bound proxy for the size of the low-rank update. 
In the main text, we therefore use
\[
    \mathcal{L}_{\mathrm{stab}}
    =
    \sum_l
    \sum_{*\in\{Q,K\}}
    \|B^l_*\|_F\|A^l_*\|_F
\]
as an adapter-stability term.

This term should not be interpreted as a direct attention loss. 
It does not compare clean and perturbed attention matrices. 
Instead, it limits one controllable source of perturbation-amplified attention reallocation: the growth of LoRA updates in the attention projections. 
The main robustness signal remains the segment-level semantic loss, while $\mathcal{L}_{\mathrm{stab}}$ provides a lightweight mechanism-side regularisation effect.

From a PAC-Bayesian perspective, LoRA adaptation can also be viewed as movement within a restricted adapter hypothesis space around a frozen pre-trained model. 
If a Gaussian prior is centred at zero LoRA update, then larger LoRA parameters correspond to a larger complexity term. 
We use this connection as a complexity-oriented interpretation of adapter stability, not as a standalone certification claim.

\section{LLM Segment}
\label{sec:appendixC}
In the main body of our work, we employ a computationally efficient punctuation-based method for segmenting model outputs. To validate this choice, we conducted an additional small-scale experiment using a more complex, high-cost segmentation approach powered by a pre-trained T5 model. This alternative method leverages the T5 model to perform semantic segmentation and alignment.

This experiment was conducted using the Bart-base model. The T5-based segmentation approach proved to be exceptionally resource-intensive, with a computational cost approximately 60 times higher than our standard punctuation-based method. Due to these practical constraints, we performed this validation on a smaller subset, using 1/8th of the original CNN/Dailymail and XSum datasets.

Fig.~\ref{fig:segment_loss_comparison_combined} below illustrates the learning trends of the inner-loop segment loss (the semantic shift loss $\mathcal{L}_{\text{sem}}$) for the first 600 fine-tuning steps (batch size: 32) on the CNN/DailyMail and XSum datasets, respectively.

\begin{table*}[t]
    \centering
    \resizebox{13cm}{!}{
    \small
    \begin{tabular}{llcccccccccc}
    \toprule
    \multirow{1}{*}{Model} & \multirow{1}{*}{Dataset} & \multirow{1}{*}{Method}&
    \multicolumn{1}{c}{A $F_{sum}$} & \multicolumn{1}{c}{B $F_{sum}$} & \multicolumn{1}{c}{LoRA $\Delta F_{sum}$}&\multirow{1}{*}{LoRA $Prod F_{sum}$}\\
    \midrule
    \multirow{4}{*}{Bart-Base}
    & \multirow{2}{*}{CNN/DM}
    & R         &19.275&9.784&57.515&184.472\\
   & & S         &18.542&7.831&64.595&140.354\\
   \cmidrule(lr){2-7}
    & \multirow{2}{*}{Xsum}
    & R         &17.241&9.103&49.177&154.976\\
   & & S         &15.977&6.562&57.687&100.997\\
    \midrule
    \multirow{2}{*}{T5-base}
    & \multirow{2}{*}{PubMed}
    & R           &49.139&18.435&208.239&875.407\\
   & & S         &40.838&18.282&142.747&691.655\\
    \midrule
    \multirow{2}{*}{Mistral-7B}
    & \multirow{2}{*}{PubMed}
    & R           &324.543&164.305&160.591&444.693\\
   & & S        &303.052&151.051&152.001&359.279\\
    \bottomrule
    \end{tabular}}
\caption{Frobenius norm statistics of LoRA parameters after standard fine-tuning (without S$^2$R$^2$). Methods R and S correspond to the R3F and SMART baselines, respectively.}
\label{tab:lora_norm_analysis}
\end{table*}

\subsection{Analysis of Results}

\begin{figure}[t]
    \centering
    \begin{subfigure}[b]{0.45\linewidth}
        \centering
        \includegraphics[width=\linewidth]{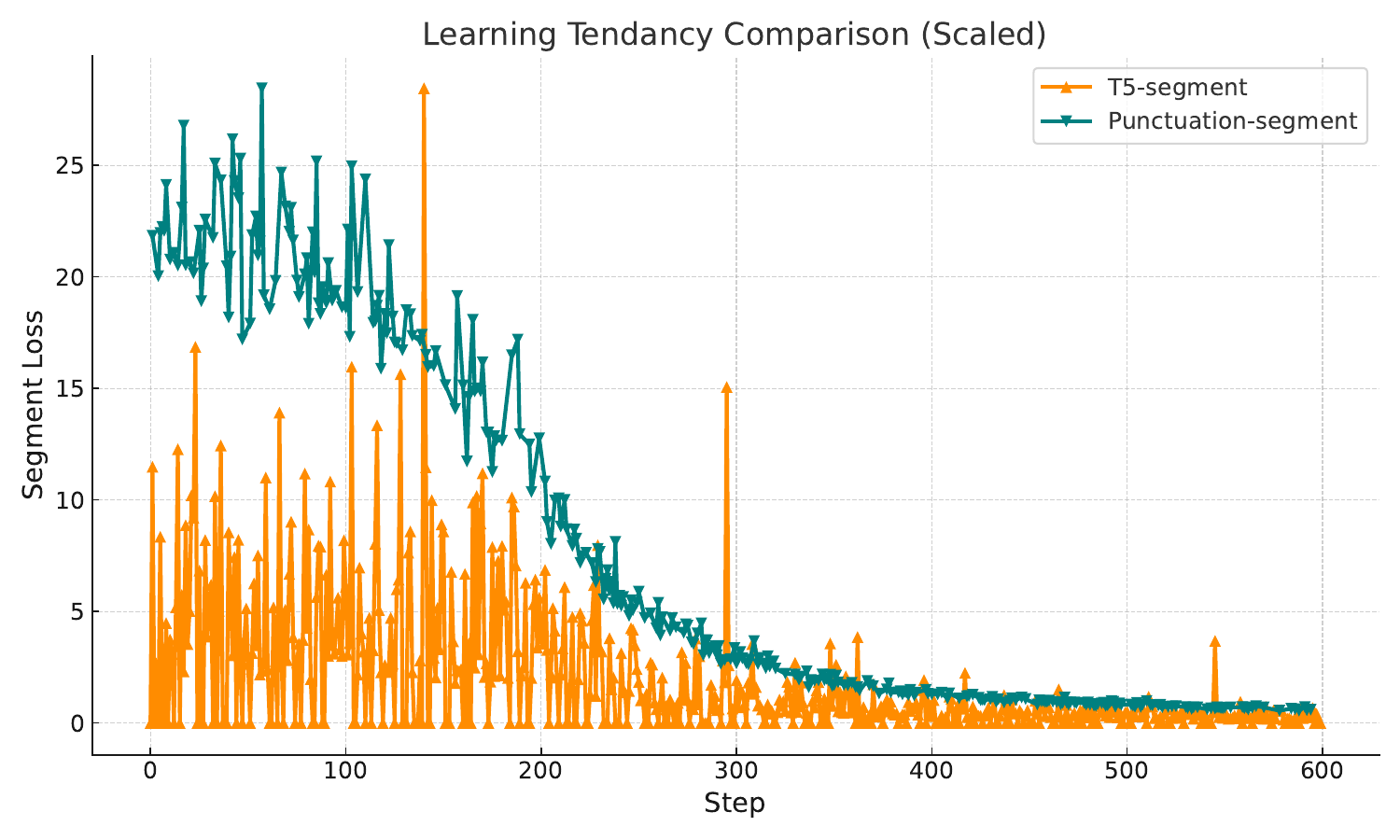}
        \caption{CNN/DailyMail Dataset}
        \label{fig:cnn_comparison_sub}
    \end{subfigure}
    \begin{subfigure}[b]{0.45\linewidth}
        \centering
        \includegraphics[width=\linewidth]{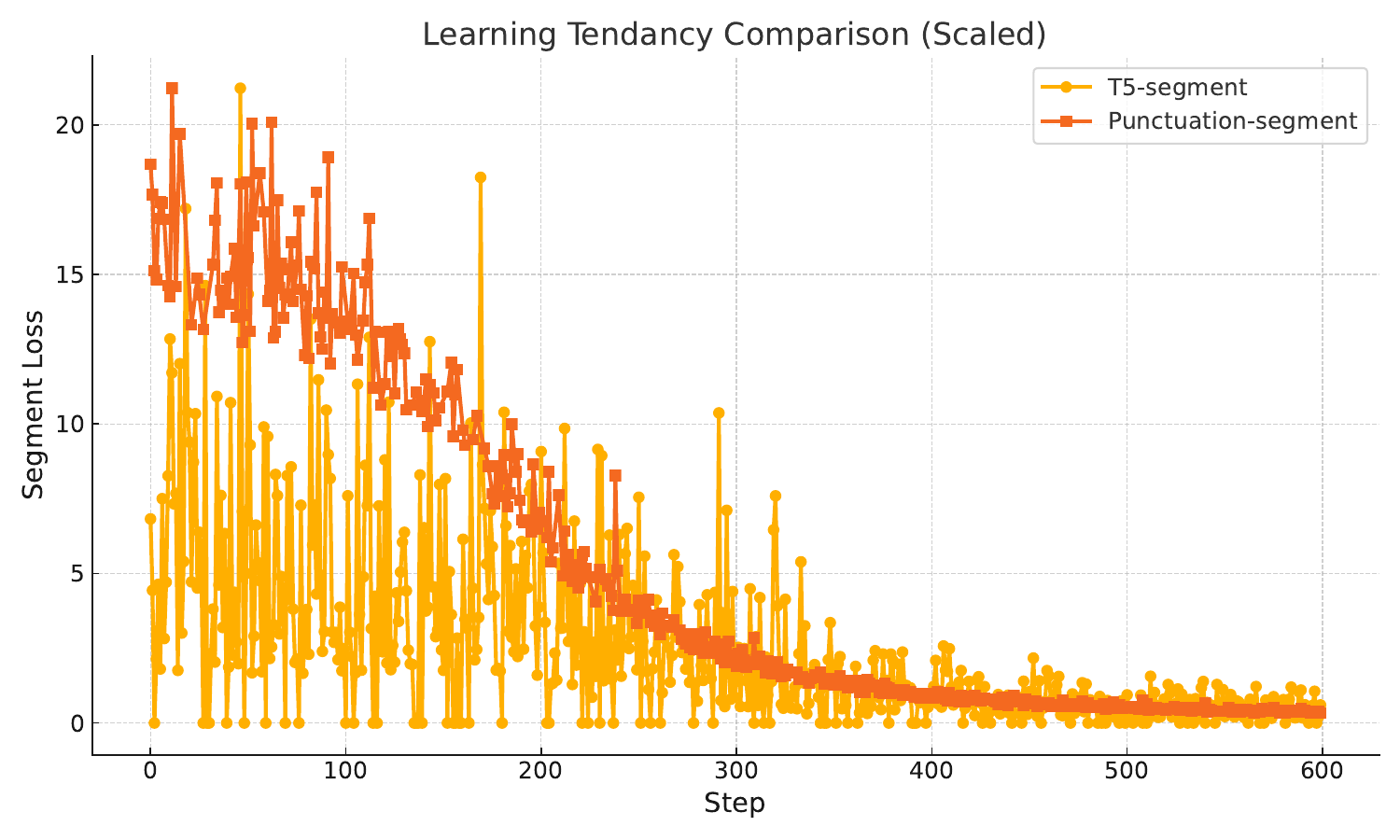}
        \caption{XSum Dataset}
        \label{fig:xsum_comparison_sub}
    \end{subfigure}
    \caption{Comparison of segment-loss learning trends for \textsc{bart-base} during the first 600 fine-tuning steps. Both punctuation-based and T5-based segmentation lead to decreasing segment loss, suggesting that the segment-level objective is optimisable under different segmentation strategies.}
    \label{fig:segment_loss_comparison_combined}
\end{figure}

From the comparison plots (Fig.~\ref{fig:segment_loss_comparison_combined}), we can draw two key observations:
\begin{enumerate}
    \item \textbf{General Convergence:} Both segmentation methods demonstrate a clear downward trend in segment loss on both datasets. This indicates that both the high-cost T5-based method and the efficient punctuation-based method are viable strategies, successfully guiding the model to minimise the semantic discrepancy between outputs from clean and perturbed inputs.

    \item \textbf{Training Dynamics and Sensitivity:} A notable difference emerges in the dynamic characteristics of the two loss curves. While both converge, the punctuation-based approach yields a smoother loss curve, whereas the signal from the T5-based segmentation is more volatile. We interpret this volatility not as training instability, but as an indicator of higher sensitivity. This characteristic likely stems from two aspects: first, the inherent complexity introduced by using a large pre-trained model as a segmentation tool; second, and more importantly, the finer granularity of the segmentation itself. By identifying more detailed semantic units, the T5-based method enables the loss function to more acutely capture the maximum distance between misaligned fragments. 
   These results suggest that the efficient punctuation-based segmentation provides a usable optimisation signal for the main experiments. 
The T5-based segmentation is more fine-grained and more sensitive, but its substantially higher cost makes it less suitable for full-scale fine-tuning. 
We therefore use punctuation-based segmentation in the main experiments and treat the T5-based results as a robustness check of the segment-level formulation.
\end{enumerate}

\section{Lora Assumption Validation}
\label{sec:appendixElora}

To validate our Assumption 2 regarding LoRA parameters as stated in the main paper, this section provides an analysis of the variance in their Frobenius norms over the course of a standard fine-tuning process (i.e., without our proposed S$^2$R$^2$). Assumption 2 posits that the Frobenius norms of the LoRA matrices, $||\bm{A}^{l}||_{F}$ and $||\bm{B}^{l}||_{F}$, are comparable.

Tab.~\ref{tab:lora_norm_analysis} presents the aggregated Frobenius norm statistics for various models after fine-tuning on their respective datasets. The metrics are defined as follows:
\begin{itemize}
    \item \textbf{A $\boldsymbol{F_{sum}}$}: The sum of the Frobenius norms of matrix A across all LoRA layers, i.e., $\sum_{l} ||\bm{A}^{l}||_F$.
    \item \textbf{B $\boldsymbol{F_{sum}}$}: The sum of the Frobenius norms of matrix B across all LoRA layers, i.e., $\sum_{l} ||\bm{B}^{l}||_F$.
    \item \textbf{LoRA $\boldsymbol{\Delta F_{sum}}$}: The sum of the absolute differences between the norms of matrices A and B for each layer, i.e., $\sum_{l} | ||\bm{A}^{l}||_F - ||\bm{B}^{l}||_F |$. This metric measures the symmetry or balance in the magnitudes of the LoRA matrices at each layer.
    \item \textbf{LoRA $\boldsymbol{Prod F_{sum}}$}: The sum of the products of the norms of matrices A and B for each layer, i.e., $\sum_{l} (||\bm{A}^{l}||_F \cdot ||\bm{B}^{l}||_F)$.
\end{itemize}

\subsection{Analysis of Norm Comparability}

By examining the values of A~$F_{sum}$ and B~$F_{sum}$ in Tab.~\ref{tab:lora_norm_analysis}, we can empirically assess the validity of Assumption 2. The data reveals a consistent trend across all models, datasets, and baseline methods: while the norms of matrices A and B are not identical, they consistently remain within the same order of magnitude.

For instance, with the \texttt{Bart-Base} model, the ratio of A~$F_{sum}$ to B~$F_{sum}$ is approximately 2.0-2.4. For the larger \texttt{T5-base} and \texttt{Mistral-7B} models, this ratio remains in a similar range, approximately 2.0-2.7. In the context of neural network parameter magnitudes, a difference of a factor of 2-3 is generally considered comparable, especially when contrasted with scenarios where parameters might differ by several orders of magnitude. This observation indicates that neither matrix's norm grows disproportionately large while the other shrinks to near zero.

This empirical result aligns with the discussion in our main paper, which acknowledges the potential for asymmetry in the roles of matrices A and B while maintaining that their magnitudes are often observed to be comparable. This empirical observation supports our use of the product-form adapter-stability term as a practical proxy for LoRA movement. 
It also makes the PAC-Bayesian complexity interpretation more plausible, since the product-form term remains connected to the quadratic norm quantity when the two LoRA factors are not extremely imbalanced.

\section{Large Language Model Usage Statement}
\label{sec:appllmuse}
During the preparation of this manuscript, we utilised an LLM, specifically OpenAI's GPT-5, to assist with language editing and polishing. The primary uses of the LLM were for improving grammar, spelling, clarity, and overall readability.

We wish to clarify that all core scientific contributions, including the conceptualisation of ideas, the design of the methodology, the execution of experiments, and the interpretation of results, are entirely the work of the human authors. The LLM served exclusively as a writing aid and did not contribute to any of the substantive research aspects of this paper. The authors have carefully reviewed and edited all text generated or modified by the LLM and take full responsibility for the final content and its scientific accuracy.

\section{Pseudocode for S$^2$R$^2$ Framework}
\label{app:pseudocode}

\begin{algorithm}[H]
\caption{Training procedure of S$^2$R$^2$}
\label{alg:s2r2}
\begin{algorithmic}[1]
\Require Pre-trained model $f(\cdot;W_0,\theta)$; training dataset $\mathcal{D}$; initial LoRA parameters $\theta_0$; perturbation set $\mathcal{P}(x)$; learning rate $\alpha$; loss weights $\lambda_{\mathrm{sem}}$ and $\lambda_{\mathrm{stab}}$.
\For{$k=1,\ldots,K$}
    \State Sample $(x,y)\sim\mathcal{D}$.
    \State Generate or sample a perturbed input $x'\in\mathcal{P}(x)$.
    \State Compute the supervised loss $\mathcal{L}_{\mathrm{CE}}(\theta_{k-1};x,y)$.
    \State Generate clean and perturbed outputs, and compute the segment-level semantic loss $\mathcal{L}_{\mathrm{sem}}(\theta_{k-1};x,x')$.
    \State Compute the adapter-stability term $\mathcal{L}_{\mathrm{stab}}(\theta_{k-1})$ from the LoRA matrices.
    \State Form the total objective:
    \[
        \mathcal{L}_{\mathrm{S^2R^2}}
        =
        \mathcal{L}_{\mathrm{CE}}
        +
        \lambda_{\mathrm{sem}}\mathcal{L}_{\mathrm{sem}}
        +
        \lambda_{\mathrm{stab}}\mathcal{L}_{\mathrm{stab}} .
    \]
    \State Update LoRA parameters:
    \[
        \theta_k
        \leftarrow
        \theta_{k-1}
        -
        \alpha\nabla_{\theta}
        \mathcal{L}_{\mathrm{S^2R^2}} .
    \]
\EndFor
\State \Return $\theta_K$.
\end{algorithmic}
\end{algorithm}

\end{document}